\documentclass{llncs}
\usepackage{graphicx, amsmath, stmaryrd, amssymb, subfig, bbm}
\usepackage[ruled,vlined]{algorithm2e}
\usepackage{url}

\newtheorem{hypothesis}{H}

\begin{document}

\title{Quiet in Class : Classification, Noise and the Dendritic Cell Algorithm}
\author{Feng Gu, Jan Feyereisl, Robert Oates, Jenna Reps, \\Julie Greensmith and Uwe Aickelin}
\institute{School of Computer Science, University of Nottingham\\ Nottingham, NG8 1BB, UK \\ \texttt{fxg@cs.nott.ac.uk}} 

\maketitle

\begin{abstract}
Theoretical analyses of the Dendritic Cell Algorithm (DCA) have yielded several criticisms about its underlying structure and operation.  As a result, several alterations and fixes have been suggested in the literature to correct for these findings.  A contribution of this work is to investigate the effects of replacing the classification stage of the DCA (which is known to be flawed) with a traditional machine learning technique.  This work goes on to question the merits of those unique properties of the DCA that are yet to be thoroughly analysed.  If none of these properties can be found to have a benefit over traditional approaches, then ``fixing'' the DCA is arguably less efficient than simply creating a new algorithm.  This work examines the dynamic filtering property of the DCA and questions the utility of this unique feature for the anomaly detection problem.  It is found that this feature, while advantageous for noisy, time-ordered classification, is not as useful as a traditional static filter for processing a synthetic dataset. It is concluded that there are still unique features of the DCA left to investigate. Areas that may be of benefit to the Artificial Immune Systems community are suggested.  
\end{abstract}

\section{Introduction}
The Dendritic Cell Algorithm (DCA) is an immune-inspired algorithm developed as part of the~\emph{Danger Project}~\cite{danger2010}.  Despite being applied to a number of applications, it was originally designed and used as an anomaly detection and attribution algorithm~\cite{greensmith2007}. For the duration of this work, the anomaly detection problem is defined as a binary classification problem, performed on (potentially noisy) discrete time series data.  The authors make no assumptions about the relative persistence of anomalous states and normal states, though the persistence of both states is assumed to be sufficiently long to differentiate them from noise.  It is also assumed that examples of a system's anomalous behaviour are available for use as training data.  This is in contrast to the many alternate definitions of the anomaly detection problem, where there can be the implicit assumption that anomalies are transient or the assumption that only normal behaviour can be studied a priori, reducing the problem to a single class classification.  For this investigation a separate, related problem is also defined, termed ``the anomaly attribution problem".  This is the problem of attributing causal relationships between the presence of elements in the environment and the occurrence of identified anomalies.   

Since its first version~\cite{greensmith2007} the DCA has been subject to many modifications~\cite{chelly2010,oates2010a}, empirical tests~\cite{alhammadi2010,greensmith2007,oates2010a} and theoretical analyses~\cite{gu2009b,musselle2010,oates2008a,stibor2009}.  This body of work has identified several interesting properties of the DCA.  For example, it has been shown that the structure of a single dendritic cell within the DCA is similar in function to the operation of a filter with a dynamically changing transfer function~\cite{oates2008a,oates2008}.  This property could be potentially useful as it allows the algorithm to both exploit the temporal ordering of the input data and remove noisy artefacts from the environmental measurements.  However, the effects of the dynamic filter within the DCA to the anomaly detection problem, beneficial or otherwise, have never been demonstrated. 

Other theoretical work identifies properties of the DCA that are clearly detrimental to its application to certain problems. One such property is that its classification stage is functionally equivalent to a statically weighted, linear classifier~\cite{stibor2009}.  Such a classifier is neither able to adapt to training data nor meaningfully act on problems which are not linearly separable.  Such a criticism is a severe blow to the utility of the DCA in its current form but only strikes at one aspect of a multifaceted algorithm.  Within the literature, it has been suggested that replacing the classification stage of the DCA with a trainable, nonlinear, machine learning algorithm would negate much of the criticism made of the DCA while preserving its novel properties~\cite{guzella2008,oates2010,stibor2009}.  

Modifying the DCA to compensate for the weaknesses identified within the literature, while retaining its original properties, is only a valid course of action if those properties are clearly beneficial.  In summary, it is important to identify if the overhead of ``fixing'' the DCA carries sufficient benefit over creating a new technique for solving the anomaly detection problem. This work is a step towards validating the usefulness of the DCA's novel properties by separating the algorithm into its component parts and assessing their individual contributions.  The structure of the paper is as follows, Section 2 provides an outline of the related work; Section 3 gives the research aims in the form of hypotheses; Section 4 presents algorithmic details as mathematical functions; Section 5 details the experimental design; Section 6 shows the results of conducted experiments and the corresponding analysis; finally Section 7 is a discussion of the findings and highlights the future steps for this work.  

\section{Related Work}
\subsection{The Dendritic Cell Algorithm}
Several different versions of the DCA exist within the literature. The deterministic DCA (dDCA) that was developed for ease of analysis, will be the version considered in this work. The algorithmic details can be found in \cite{greensmith2008b}.

The first stage of the DCA is an anomaly detection phase, where the population's classification decisions are monitored in order to identify anomalies within a given dataset. The second phase attempts to correlate the antigen sampled by the cells with the occurrence of detected anomalies.

The DCA receives two types of input, namely signal and antigen. Signals are represented as vectors of real-valued numbers and are periodic measurements of features within the problem environment. An assumption made by the algorithm is that the presence or absence of an anomaly can be detected by observing these features. Antigen are symbols (typically represented as an enumerated type), which represent items of interest within the environment. It is assumed that some of the antigen have a causal relationship with observed anomalies.

The DCA is a population-based algorithm, where several heterogenous agents (cells) monitor the same inputs in parallel. Each cell stores a history of the received input signals, while maintaining a sum of their magnitudes. Upon the sum of the input signal magnitudes reaching a predefined decision threshold, the cell performs a classification based on the signal history. When the decision has been recorded, the cell is reset and instantaneously returned to the population. Each cell is assigned a different decision threshold generated from a uniform distribution, ensuring that cells observe the data over different time scales. 

It is demonstrated in~\cite{stibor2009} that both the classification boundary and the position of the decision boundary can be expressed as hyperplanes, akin to those found in linear classifiers. This premise is used as a foundation for this investigation, so the pertinent machine learning concepts are presented in Section~\ref{Sec_MLC}. As the classification performed by a cell is performed using the history of the sampled signals rather than an instantaneous sample of the environmental features, it can be shown that the DCA exhibits a filtering property which allows it to remove high frequency noise from the input signals~\cite{oates2008a}. This process relies on the underlying state of the system (normal or anomalous) being persistent for a long enough period of time to distinguish it from the noise. This filtering property is also a key premise of this work and shall be discussed in greater depth in Section~\ref{Sec_SPC}.

\subsection{Machine Learning Concepts}\label{Sec_MLC}
In our investigation, the classification stage of the DCA is replaced by a trainable classifier, which is based on the operation of Support Vector Machines (SVM)~\cite{burges1998}. Here we present an introduction to this algorithm and the relevant machine learning concepts. SVM models can be described using linear discriminant functions~\cite{duda2000}, quadratic optimisation~\cite{fletcher1987}, and kernel methods~\cite{scholkopf2002}.

Let $(\mathbf{x}_{1},y_{1}), ... , (\mathbf{x}_{n},y_{n})\in\mathbf{X}\times Y$ be a given training set with $n$ data instances, where $\mathbf{X}\subseteq\mathbb{R}^{d}$ is a $d$-dimensional input feature space and $Y=\{\pm1\}$ is a set of truths or class labels. For each data instance $\mathbf{x}_{i}\in\mathbf{X}$ where $i\in[1,n]\cap\mathbb{N}$, a linear discriminant function is defined as $f : \mathbb{R}^{d}\rightarrow\mathbb{R}$,
\begin{equation}
f(\mathbf{x}_{i})=\langle\mathbf{w},\mathbf{x}_{i}\rangle+b
\end{equation}
where $\langle\cdot\rangle$ denotes the inner product of two vectors, $\mathbf{w}$ is the weight vector and $b$ is the bias. The decision boundary of classification is given by $\langle\mathbf{w},\mathbf{x}\rangle+b = 0$, which corresponds to a $(d-1)$-dimensional hyperplane within a $d$-dimensional feature space. A signed measure of the perpendicular distance $r$ from the decision surface to a data point $\mathbf{x}$ can be calculated as,
\begin{equation}
r = \frac{f(\mathbf{x})}{\|\mathbf{w}\|}
\label{eqn:lnperdist}
\end{equation}
where $\|\cdot\|$ is the norm operator of a vector.

The linear discriminant functions of SVM models are based on the maximal margin classifier, defined as follows: 
\begin{equation}
\langle\mathbf{w},\mathbf{x}_{i}\rangle+b\ge +1\quad\mathrm{if}~y_{i}=+1 
\label{eqn:svmpos} 
\end{equation}
\begin{equation}
\langle\mathbf{w},\mathbf{x}_{i}\rangle+b\le -1\quad\mathrm{if}~y_{i}=-1 
\label{eqn:svmneg}
\end{equation}

Data points lying on the hyperplane $H_{1}: \langle\mathbf{w},\mathbf{x}\rangle+b=1$ have a perpendicular distance from the origin $|1-b|/\|\mathbf{w}\|$. Similarly, data points lying on the hyperplane $H_{2}: \langle\mathbf{w},\mathbf{x}\rangle+b=-1$ have a perpendicular distance from the origin $|-1-b|/\|\mathbf{w}\|$. The margin between the two hyperplanes $H_{1}$ and $H_{2}$ is equal to $2/\|\mathbf{w}\|$. An optimal decision boundary, defined by $\langle\mathbf{w},\mathbf{x}\rangle+b=0$, is found by maximising this margin. It is equidistant and parallel to $H_{1}$ and $H_{2}$. 

The learning task of SVM can be defined as an optimisation problem,
\begin{equation}
\left \{\begin{array}{l l}
\mathrm{minimise}_{\mathbf{w},b} &\quad\|\mathbf{w}\|^{2} \\
\mathrm{subject~to} &\quad y_{i}(\langle\mathbf{w},\mathbf{x}_{i}\rangle+b) - 1\ge 0\quad\forall i
 \end{array} \right. \\
\label{eqn:svmcom}
\end{equation}
where the constraints are derived from combining Equation~\ref{eqn:svmpos} and Equation~\ref{eqn:svmneg}. Such an optimisation problem becomes much easier to solve if we introduce Lagrangian multipliers. Let $\alpha_{i}\ge 0$ be the Lagrangian multipliers, which correspond to the constraints in Equation~\ref{eqn:svmcom}. A primal Lagrangian of the above optimisation problem is defined as
\begin{equation}
L_{P} = \frac{1}{2}\|\mathbf{w}\|^{2} - \sum_{i=1}^{n}\alpha_{i}[y_{i}(\langle\mathbf{w},\mathbf{x}_{i}\rangle+b) - 1] 
\label{eqn:svmlgprim}
\end{equation}

The primal form $L_{P}$ is differentiable with respect to $\mathbf{w}$ and $b$, and an equivalent dual form, known as the Wolfe dual~\cite{fletcher1987}, can be derived. The optimisation becomes a convex quadratic programming problem, and all data points that satisfy the constraints also form a convex set~\cite{burges1998}. This dual form is defined as
\begin{equation}
L_{D} = \sum_{i=1}^{n}\alpha_{i} - \frac{1}{2}\sum_{i,j=1}^{n}\alpha_{i}\alpha_{j}y_{i}y_{j}\langle\mathbf{x}_{i},\mathbf{x}_{j}\rangle
\label{eqn:svmlgdual}
\end{equation}

During the training phase of SVM, $L_{D}$ is maximised with respect to all $\alpha_{i}$. The solution of~\ref{eqn:svmlgdual} contains feature vectors $\mathbf{x}_{i}$ such that their corresponding $\alpha_{i}\neq0$. These vectors are called support vectors, and they lie on either $H_{1}$ or $H_{2}$. For non-separable cases, additional constraints are required to allow for outliers. These constraints are $\sum y_{i}\alpha_{i}=0$ and $0\le\alpha_{i}\le C$, where $C$ is a parameter that controls the regularisation term. In addition, $\langle\mathbf{x}_{i},\mathbf{x}_{j}\rangle$ can be replaced by $\langle\Phi(\mathbf{x}_{i}),\Phi(\mathbf{x}_{j})\rangle$ through kernel methods. A kernel function is defined as,
\begin{equation}
k(\mathbf{x}_{i}, \mathbf{x}_{j}) = \langle\Phi(\mathbf{x}_{i}),\Phi(\mathbf{x}_{j})\rangle
\label{eqn:kernel}
\end{equation}
where $\Phi$ is a mapping from the original input feature space $\mathbf{X}$ to a higher dimensional (inner product) feature space $F$, where nonlinearly separable problems become more separable~\cite{scholkopf2002}.  

Depending on the applications, a number of kernel functions are available, including linear kernels, polynomial kernels, and Gaussian kernels~\cite{scholkopf2002}. A linear kernel only involves performing inner product operations with the input data. Therefore a linear SVM that uses such a kernel is usually simple to train and use. It is more computationally efficient than other SVM models that use more complicated kernel functions~\cite{fu2010}. The linear SVM is chosen in this work due to its algorithmic and computational simplicity. 

\subsection{Signal Processing Concepts}\label{Sec_SPC} 
Filters can be viewed as algorithms or structures which apply a gain (ratio of output to input), to their input signal to produce a new output signal. Where filters differ from a simple amplifier, is that the gain applied is a function of the frequency of the input. The mathematical function relating gain to frequency is referred to as the ``transfer function'' of the filter. In the field of signal processing it is common practice to express filters by providing their transfer functions. For completeness the filters being used for this work will be given here.

The filter with the most analogous behaviour to the DCA is the sliding window filter~\cite{oates2008a}. A sliding window filter is so called as it can be viewed as a bounding box being translated along the input data. At each step $t$, the output of the sliding window filter is the average sample size contained within the window. This is expressed in Equation~\ref{SLW_eq},
\begin{equation}
\label{SLW_eq}
o_t = \frac{1}{W}\sum_{a=(t-W)}^{t}i_a
\end{equation}
where $o_t$ is the output of the filter at step $t$, $i_a$ is the input sample at time index $a$ and $W$ is the width of the window in steps.

The transfer function of the sliding window filter is given in Equation~\ref{SLTF_eq}~\cite{ifeachor2001},
\begin{equation}
\label{SLTF_eq}
G_{S}(\omega) = \frac{1}{W}\sum_{g=0}^{W-1}e^{-jg\omega}
\end{equation}
where $G_{S}(\omega)$ is the transfer function of the sliding window filter, $j$ is the complex number constant and $\omega$ is the frequency of the input signal.

A dendritic cell acts like a sliding window filter which only reports its output every $W$ steps~\cite{oates2008a}. The transfer function for such a filter is given in Equation~\ref{DCTF_eq},
\begin{equation} \label{DCTF_eq} 
G_{D}(\omega) = \frac{1}{W^2}\sum_{g=0}^{W - 1}\sum_{b=0}^{W -1}e^{-jb((\omega+(2g\pi)))} 
\end{equation}
where $G_{D}(\omega)$ is the transfer function of the dendritic cell. However, this transfer function assumes a constant window size $W$. For a dendritic cell the window size is a function of the magnitude of the input signal being filtered and the decision boundary assigned to the cell. This makes expressing a cell's transfer function extremely difficult as the magnitude of the signal cannot be known a priori. With a given training set, a suitable value to use for the decision boundary could be found by minimising the classification error. However, it is not known if this dynamically changing window size is of any benefit to the algorithm.

\section{Research Aims}\label{sec_RA}
To justify future work on the DCA it is necessary to assess the importance of its novel features. In the literature, three novel properties of the DCA remain unvalidated: the effect of antigen; the effect of the dynamic filtering; and the effect of having a population of classifiers. Of these, it is arguable that the effect of the dynamic filtering is the most important. This is because the antigen effect is unlikely to yield positive results if the anomaly detection phase is insufficient and the classifier population is unlikely to yield positive results if the dynamic filters used by that population prove to be insufficient.  

In order to verify the need for a filter of any kind, it is important to determine if filtering the output from a classifier improves the results of classification when using a time-ordered, noisy dataset. The following null hypothesis will be the first step in this investigation.

\begin{hypothesis}
\begin{minipage}[t]{3.8in}
Filtering the results of a linear classifier presented with time ordered, noisy input data will not result in significant difference of the classification performance.
\end{minipage}
\end{hypothesis}

This is obviously dependent on designing an appropriate filter as part of the experimental setup.

In order to justify the additional implementation complexity, the dynamic filters should outperform a suitably tuned static counterpart.  This yields the following testable null hypothesis.

\begin{hypothesis}
\begin{minipage}[t]{3.8in}
The results from a linear classifier filtered by a dynamic moving window function will have no significant difference to the results from the same classifier using a static moving window function.
\end{minipage} 
\end{hypothesis}

While this investigation is not primarily focussed on the other novel features of the DCA it is of interest to compare the output from the original DCA to that of a filtered and an unfiltered classifier.  A trained classifier may have the advantage of being able to adapt to the input data, but the DCA has the additional antigen and multiple perspectives properties, so it will be difficult to definitively identify the reasons for relative performance. However, should the DCA outperform a filtered classifier, it shows that the other properties of the DCA add some information to the decision making process.  If on the other hand the DCA is outperformed by a filtered classifier, it would suggest that the benefits of adding a training phase, at the very least, outweigh the possible benefits of the other novel aspects of the algorithm.  In either case more experiments would need to be done to assess the merits of the other algorithmic properties.  The testable hypothesis from this investigation's perspective is as follows.

\begin{hypothesis}
\begin{minipage}[t]{3.8in}
The classification performance of the DCA will not be significantly different to that of a linear classifier, filtered or otherwise on a time-ordered, noisy dataset. 
\end{minipage}
\end{hypothesis}

If it is possible to reject all of these null hypotheses, then a second set of statistical tests can be performed, assessing the relative benefit of using one technique over the other for the dataset used.

\section{Algorithmic Details} \label{sec_AD}
To investigate the merits of the sliding window effect of the DCA, it is necessary to separate it from the rest of the algorithm, and use it in conjunction with a better understood classifier. For this investigation, two moving window functions are used as filters for processing the decisions made by a linear SVM. For a given training set, the linear SVM finds an optimal decision boundary and returns the signed orthogonal distance from the decision boundary to each data point, as defined in Equation~\ref{eqn:lnperdist}. The moving window functions initialise either a set of window sizes or a set of decision thresholds, and label the data instances within every moving window created.  An error function is used to find the optimal window size or decision threshold. The knowledge obtained through training is then applied to classify data instances within the testing set. 

For clarity, the algorithmic combinations of a linear SVM with a static and dynamic moving window function that are used in the experiments are defined in the subsequent sections. As the dynamic moving window function cannot be easily defined in a continuous frequency domain, we define both moving window functions in a discrete time domain. For this section time is indexed by $i\in[1,n]\cap\mathbb{N}$ i.e. the index of a data instance in the feature space.

\subsection{Static Moving Window Function}
Let $A=\{\alpha_{l}\mid\alpha_{l}\in\mathbb{N}\}$ be a set of $m$ initial window sizes where $l\in[1,m]\cap\mathbb{N}$, and $k\in[1,\lceil\frac{n}{\alpha_{l}}\rceil]\cap\mathbb{N}$ be the index of a moving window depending on $\alpha_{l}$, where $\lceil\cdot\rceil$ denotes the ceiling function. Let $S_{k}=[1+(k-1)\alpha_{l},k\alpha_{l}]\cap\mathbb{N}$ be a set of indexes of the data instances contained within a static moving window. This divides the entire interval $[1,n]\cap\mathbb{N}$ into $\lceil\frac{n}{\alpha_{l}}\rceil$ partitions. The function for determining the class label of each data instance with respect to a window size $\alpha_{l}$ is defined as $c : \mathbb{R}^{n}\times\mathbb{N}\times\mathbb{N}\rightarrow\{\pm1\}$,
\begin{equation}
c(f(\mathbf{x}),\alpha_{l},i)=\sum_{k=1}^{\lceil\frac{n}{\alpha_{l}}\rceil}\mathbbm{1}_{S_{k}}(i)~\mathrm{sgn}\left(\sum_{s=1}^{n}\frac{f(\mathbf{x}_{s})}{\|\mathbf{w}\|}\mathbbm{1}_{S_{k}}(s)\right)
\label{eq:winlab}
\end{equation}
where $\mathbbm{1}_{X}(x)$ defines an indicator function that returns one if $x\in X$ holds and zero otherwise, and $\mathrm{sgn}(\cdot)$ denotes a sign function of real numbers defined as,
\begin{equation}
\mathrm{sgn}(x)=\left \{\begin{array}{l l}
+1 &~\mathrm{if}~x\ge0 \\
-1 &~\mathrm{otherwise} \end{array} \right.
\end{equation}
where $x\in\mathbb{R}$. For each window size $\alpha_{l}$, the function $c$ firstly calculates the cumulative distance of all data points, within a generated window, with respect to the decision boundary. It then labels each data instance within such window according to the sign of the calculated cumulative distance. This process is iterative for all the windows generated with respect to a window size. 

A mean square error based function is used for evaluating the effectiveness of each window size with respect to the class label, defined as $e : \mathbb{N}\rightarrow\mathbb{R}$.
\begin{equation}
e(\alpha_{l})=\frac{1}{n}\sum_{i=1}^{n}\|c(f(\mathbf{x}),\alpha_{l},i) - y_{i}\|^{2}
\label{eq:erates}
\end{equation}
The static moving window function returns an optimal window size $\alpha_{opt}\in A$ that minimises the resulting classification error, defined as
\begin{equation}
\alpha_{opt}= \arg\min_{\alpha_{l}\in A}\{e(\alpha_{l})\}  
\end{equation}

\subsection{Dynamic Moving Window Function}
Let $B=\{\beta_{l}\mid \beta_{l}\in\mathbb{R}\}$ where $l\in[1,m]\cap\mathbb{N}$ be a set of $m$ initial decision thresholds (lifespans), and $k\in[1,\lfloor\frac{\sum f(\mathbf{x}_{i})}{\beta_{l}}\rfloor]\cap\mathbb{N}$ be the index of a moving window depending on the decision threshold $\beta_{l}$, where $\lfloor\cdot\rfloor$ denotes the floor function. For each decision threshold $\beta_{l}$, the moving windows are found by the following inequality,
\begin{equation}
b_{l}^{k+1}=\arg\max_{a\in\mathbb{N}}\left\{a\in(b_{l}^{k},n]\mid\sum_{i=b_{l}^{k}}^{a}\left|\frac{f(\mathbf{x}_{i})}{\|\mathbf{w}\|}\right|\le \beta_{l}\right\}\quad\forall k
\label{eq:lsbound}
\end{equation}
where $|\cdot|$ is the absolute operator, and each dynamic moving window is bounded by $[b_{l}^{k},b_{l}^{k+1}]\cap\mathbb{N}\subseteq[1,n]\cap\mathbb{N}$, where $b_{l}^{k}$ and $b_{l}^{k+1}$ are the beginning and end points of the $k$th moving window and $b_{l}^{1}=1$. The dynamic window size of a decision threshold $\beta_{l}$ is bounded by the cumulative absolute distances $|r_{i}|=|f(\mathbf{x}_{i})/\|\mathbf{w}\||$ from the optimal decision boundary to all the points within it. This is due to the magnitude of $|r_{i}|$ being closely related to the degree of confidence (sufficient information) for making a decision regarding classification. Let $\tilde{S}_{k}=[b_{l}^{k},b_{l}^{k+1}]\cap\mathbb{N}$ be a set of indexes of the data instances contained within a dynamic moving window. This divides the entire interval $[1,n]\cap\mathbb{N}$ into $\lceil\frac{\sum f(\mathbf{x}_{i})}{\beta_{l}}\rceil$ partitions. A similar function to Equation~\ref{eq:winlab} for labelling each data instance with respect to a decision threshold $\beta_{l}$ is defined as $\tilde{c} : \mathbb{R}^{n}\times\mathbb{R}\times\mathbb{N}\rightarrow\{\pm1\}$.
\begin{equation}
\tilde{c}(f(\mathbf{x}),\beta_{l},i)=\sum_{k=1}^{\lfloor\frac{\sum f(\mathbf{x}_{i})}{\beta_{l}}\rfloor}\mathbbm{1}_{\tilde{S}_{k}}(i)~\mathrm{sgn}\left(\sum_{s=1}^{n}\frac{f(\mathbf{x}_{s})}{\|\mathbf{w}\|}\mathbbm{1}_{\tilde{S}_{k}}(s)\right)
\label{eq:lslab}
\end{equation} 

Similar to Equation~\ref{eq:erates}, a mean square error based function with respect to the class label is used for assessing the effectiveness of each decision threshold, defined as $\tilde{e} : \mathbb{R}\rightarrow\mathbb{R}$.
\begin{equation}
\tilde{e}(\beta_{l})=\frac{1}{n}\sum_{i=1}^{n}\|\tilde{c}(f(\mathbf{x}),\beta_{l},i) - y_{i}\|^{2}
\label{eq:lserrates}
\end{equation}
The dynamic moving window function returns an optimal decision threshold $\beta_{opt}\in B$ that minimises the resulting classification error, defined as
\begin{equation}
\beta_{opt}=\arg\min_{\beta_{l}\in B}\{\tilde{e}(\beta_{l})\}
\end{equation}

\section{Experimental Design} \label{sec_ED}
This section details the techniques used to implement the algorithms of interest and the synthetic data required to test the null hypotheses outlined in Section~\ref{sec_RA}. Details of the raw datasets, experimental results and statistical analyses involved in this paper can be found in~\cite{icarisdat2011}. 

\subsection{Synthetic Datasets}
Synthetic datasets based on two Gaussian distributions are common practice in machine learning, as shown in~\cite{stibor2009}. This is due to the fact that varying the distance between the distributions allows for control over the separability of the data.  For the experiments in Musselle's work~\cite{musselle2010}, where the temporal nature of the data is important, the author uses a Markov chain to generate synthetic datasets, where the probability of state change dictates the relative concentrations of the normal and anomalous behaviour. 

\begin{figure}
\centering
	\subfloat[Example 1: feature space]{\includegraphics[width=0.42\textwidth]{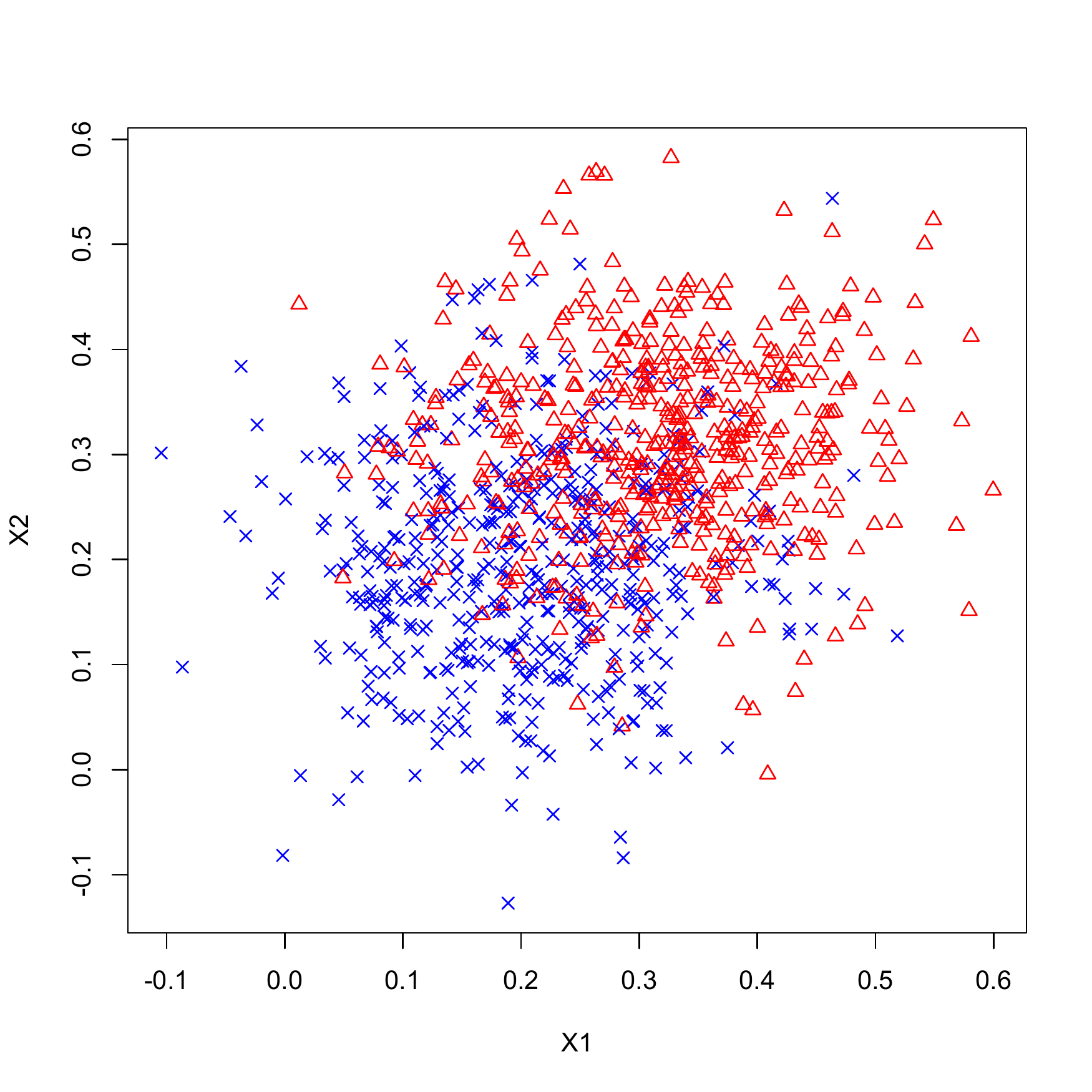}}	
	\subfloat[Example 1: time domain]{\includegraphics[width=0.42\textwidth]{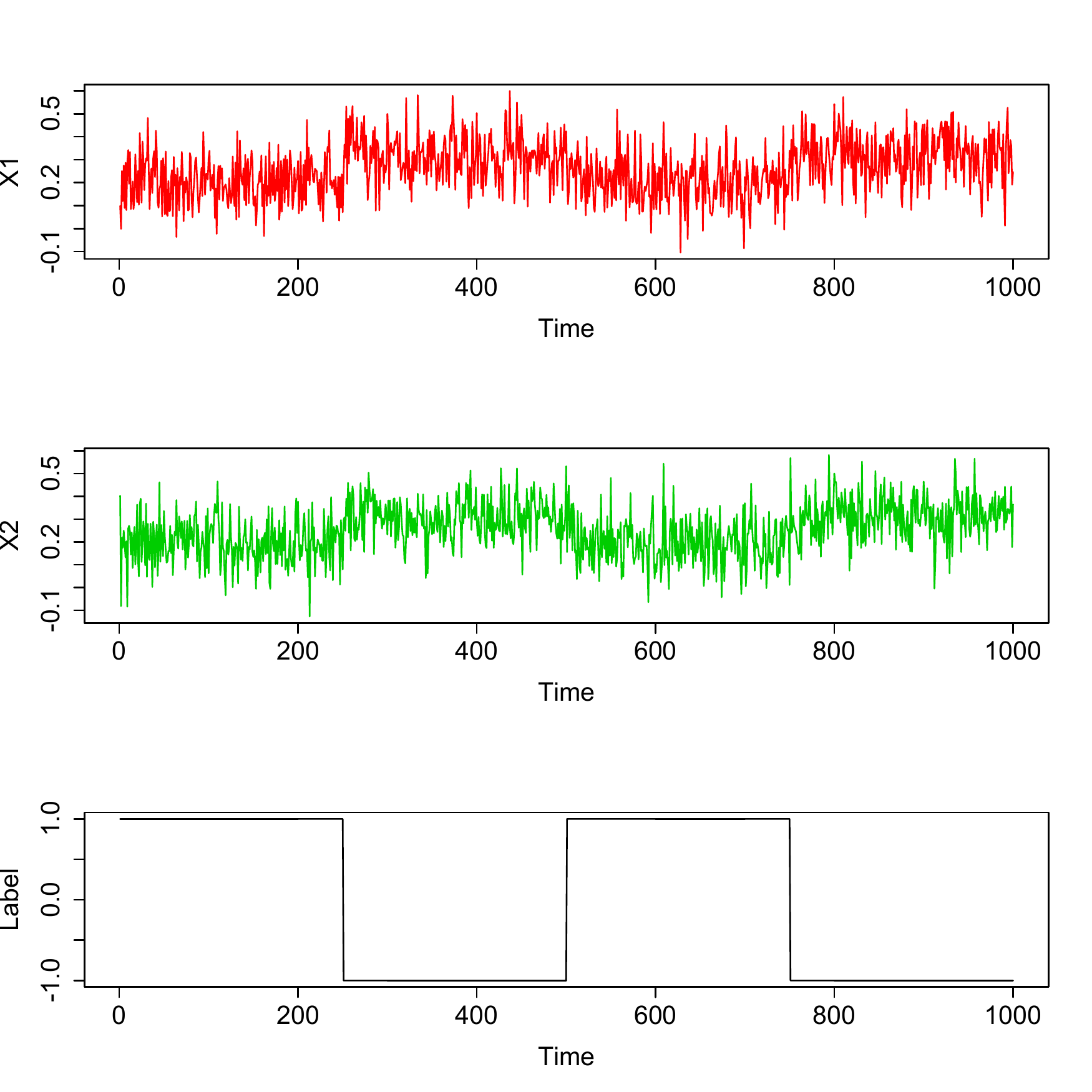}}

	\subfloat[Example 2: feature space]{\includegraphics[width=0.42\textwidth]{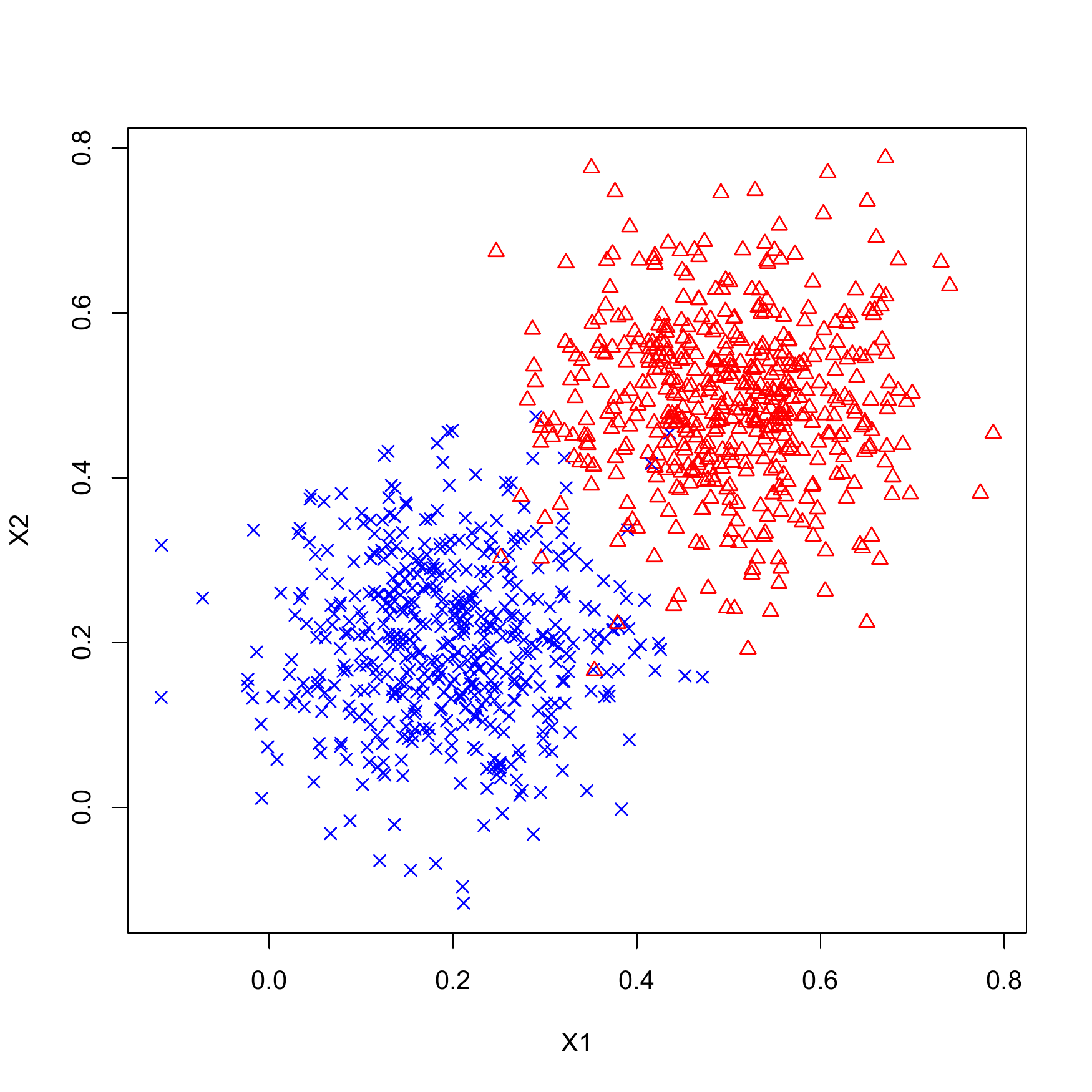}}	
	\subfloat[Example 2: time domain]{\includegraphics[width=0.42\textwidth]{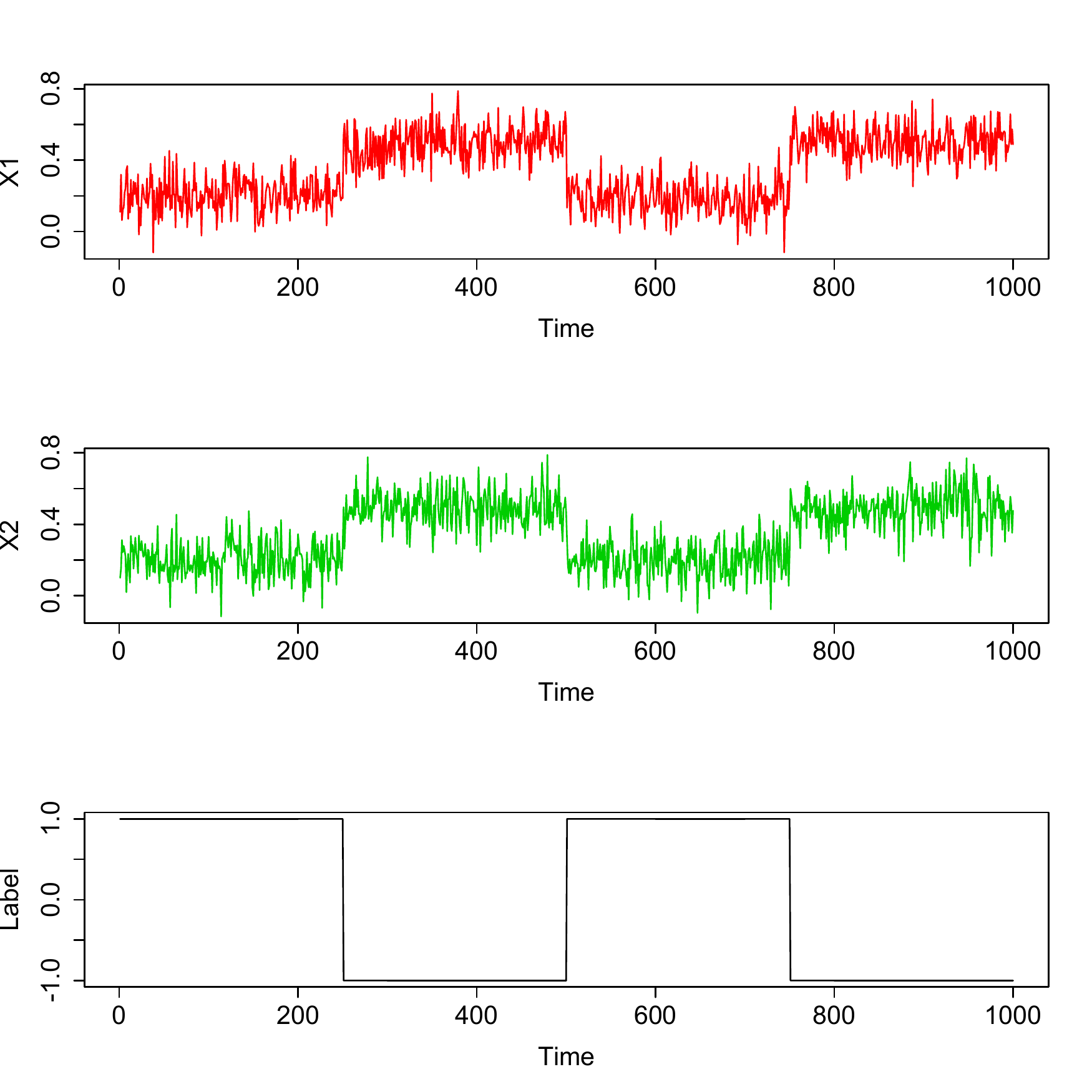}}
	
	\subfloat[Example 3: feature space]{\includegraphics[width=0.42\textwidth]{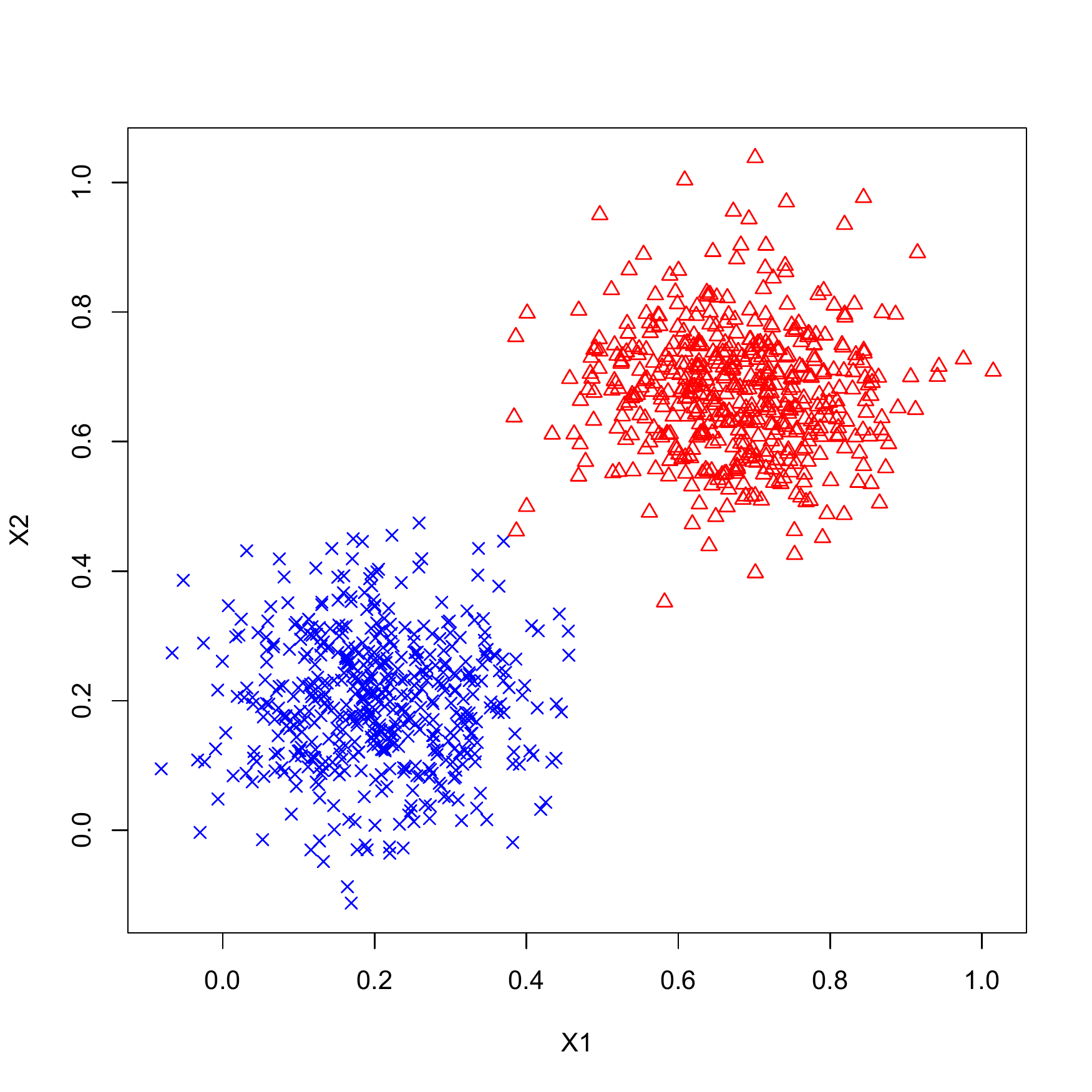}}	
	\subfloat[Example 3: time domain]{\includegraphics[width=0.42\textwidth]{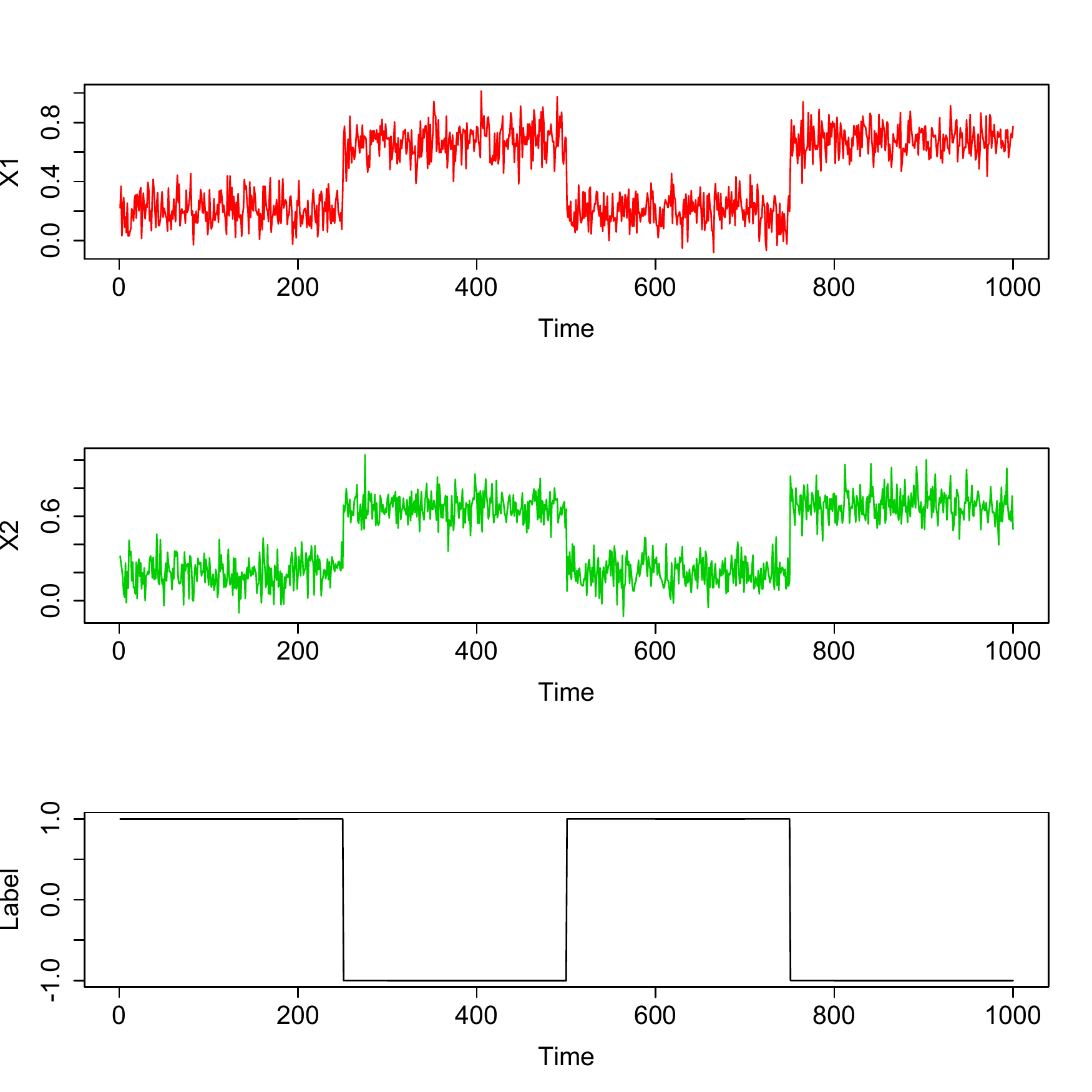}}
\caption{Feature space and time domain plots of three examples where two classes have different degrees of overlap, and the Euclidean distances between the centroids are 0.17 (a, b), 0.42 (c, d), and 0.68 (e, f).}
\label{fig:classdist}
\end{figure}

For this investigation, both separability and temporal ordering are important. Therefore it was decided to use a dataset based on two Gaussian distributions, then introduce to it an artificial temporal ordering.  This is achieved by creating time varying signals representing the class features.  Each dataset is divided into quarters, where the first and third quarters are of class I and the second and fourth quarters are of class II. This ordering provides a low frequency underlying change of class, and provides examples of class transitions in both directions.  As a consequence, by varying the separability of the classes, one also changes the signal to noise ratio of the time-ordered data, effectively maintaining the same level of noise, but increasing the magnitude of the underlying signal as the separability increases, as illustrated by Fig.~\ref{fig:classdist}. 

For the generated datasets, class I's mean is fixed at 0.2, and 100 datasets are generated by varying class II's mean from 0.2 (total overlap), to 0.8 (linearly separable) at a regular interval.  Both distributions use a standard deviation of 0.1.  As the mean of class II increases, the Euclidean distance between the centroids of the two classes increases accordingly. This corresponds to the increment in separability of the two classes. Each dataset contains 2,000 instances, 1,000 for training and 1,000 for testing.  By using large numbers of samples, it is intended to reduce artefacts caused by bias in the random number generator.  
 
\subsection{Algorithm Setup}
Parameters used in the linear SVM are the default values of the R package kernlab~\cite{karatzoglou2011}, and kept the same for both moving window functions. For the static moving window function, the cardinality of the set of initial window sizes $|A|$ is 100, and a window size $\alpha_{l}\in[1,100]\cap\mathbb{N}$. For the dynamic moving window function, the cardinality of the set of initial decision thresholds $|B|$ is also 100, and a decision threshold $\beta_{l}$ is calculated as, 
\begin{equation}
\beta_{l}=\arg\max_{\mathbf{x}_{i}\in\mathbf{X}}\left\{\frac{|f(\mathbf{x}_{i})|}{\|\mathbf{w}\|}\right\}\frac{l}{|B|}\lambda
\label{eqn:lifes}
\end{equation}
where $\lambda$ is a scaling factor that controls the window sizes considered by the parameter tuning. If $\lambda=1$, windows are constrained to values typically used within the DCA literature. With $\lambda=100$, parameters which are equivalent to those used by the static and dynamic moving window functions can also be considered by the tuning process.

The DCA often requires a preprocessing phase that is analogous to the training phase of the linear classifier algorithm, thus only testing sets are used by the DCA. Firstly the two input features are normalised into a range $[0,1]$ through min-max normalisation. The correlation coefficient between each feature and the class label is then calculated and used to map either of the features to the appropriate signal category. The remaining parameters are chosen according to the values suggested in~\cite{greensmith2008b}. The initialisation of lifespans in the DCA uses a similar principle as Equation~\ref{eqn:lifes}, however the maximisation term is replaced by the signal transformation function of the algorithm and the entire set of lifespans are used for the DC population. 

\subsection{Statistical Tests}
All results will be tested using the Shapiro-Wilk normality test to verify if parametric or nonparametric statistical tests are suitable~\cite{crawley2005}.  All of the null hypotheses in Section~\ref{sec_RA} are phrased as the absence of a detectable significant difference between pairs of results.  The two-sided student t-test will be used for normally distributed samples, and the two-sided Wilcoxon signed rank test will be used for non-normally distributed ones~\cite{crawley2005}.  

If differences are detected, the one-sided versions of the relevant difference test will be used to ascertain the relative performance of the results. For all statistical tests a significance level of $\alpha=0.05$ will be considered sufficient.

\section{Results and Analysis} \label{sec_Res}
\begin{figure}[t]
\centering
\includegraphics[width=\textwidth]{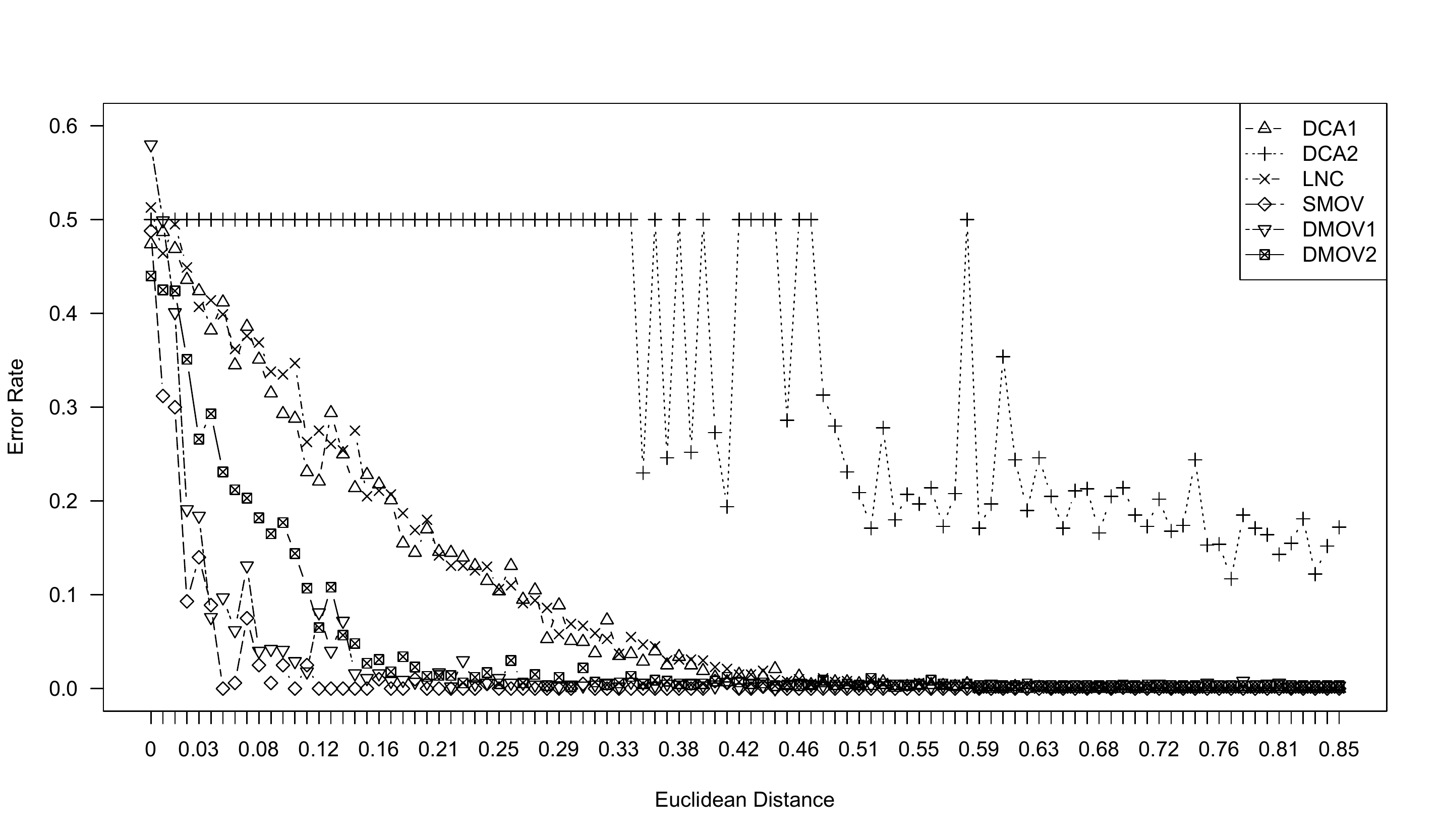}
\caption{Error rates of tested methods against the Euclidean distance between centroids of the two classes across all the datasets. DCA1 ($\lambda=1$) and DCA2 ($\lambda=100$) denote the Dendritic Cell Algorithm, LNC is the linear SVM, SMOV is the static moving window function, and DMOV1 ($\lambda=1$) and DMOV2 ($\lambda=100$) is the dynamic moving window function.}
\label{fig:erates}
\end{figure}

Results from the experiments are presented in terms of the error rates, which are equal to the number of misclassified data instances divided by the total number of instances in the tested dataset. The error rates of the six tested methods across all of the datasets are plotted against the Euclidean distance between the two class centroids in Fig.~\ref{fig:erates}. For non-separable cases, the classification performance differs from one method to another. In order to determine whether these differences are statistically significant, statistical tests are performed as follows. 

The Shapiro-Wilk tests confirm that the data are not normally distributed ({\it p}-values are less than 0.05) and therefore the Wilcoxon tests are used to assess the statistical significance for both the two-sided and one-sided comparisons described previously. As all the {\it p}-values are less than 0.05, we reject the null hypotheses of all the two-sided Wilcoxon tests with a 95\% confidence and conclude that significant differences exist between the results of the different methods. As a result, all of the three null hypotheses presented in Section~\ref{sec_RA} are rejected. 

For completeness Fig.~\ref{fig:erates} shows results for $\lambda = 1$ (the original DCA parameter range) and the extended search space of $\lambda = 100$. However for analysis, we will only consider the best performing parameterisations of each unique method. From inspection of Fig.~\ref{fig:erates}, it is argued that the order of the methods, in terms of ascending classification performance, is as follows: the linear SVM; the DCA (DCA1); the dynamic moving window function (DMOV2); and the static moving window function. As all the p-values of the one-sided Wilcoxon tests are less than 0.05, this inspection is statistically verified.

\section{Discussion and Future Work}
The experimental results demonstrate that filtering the decisions of a linear classifier presented with time-ordered and noisy input data significantly changes and improves its classification performance.  This was expected to be the case, as even when the datasets are non-separable in the feature space, the temporal ordering means that so long as the hyperplane has a greater than 50\% accuracy, it is likely that the average of several instances from the same class, will tend towards the correct class label. This can also be viewed from the frequency domain as non-separability introducing a high frequency noise component into the signal, which can be removed by filtering.  

The classification performance of the DCA is significantly different from a linear classifier, filtered or otherwise, on a time-ordered and noisy dataset. In fact, the DCA produces significantly better classification performance than a standard linear classifier, but significantly worse classification performance than the filtered linear classifiers.  This implies that the filtering property of the DCA is an important factor of its performance, but that the addition of a training phase to the DCA can add further, substantial improvements. 

It is also shown that the classification performance of a linear classifier with a static moving window function is significantly different and better, in comparison to that of a linear classifier with a dynamic moving window function. This is only a valid statement for the datasets used, but infers that the heuristic used by the DCA to alter the transfer function of its filtering component, (i.e. the magnitude of the input signal) is not as good as a simple, static filter.

These results suggest that the problems with the DCA are more deep-rooted than having linear decision boundaries.  The DCA's main advantage over the SVM seems to have been its novel filtering technique.  However, by substituting the individual components of the DCA with traditional techniques from the domains of signal processing and machine learning, it is clear that it is outperformed.  Finding equivalence between the DCA's properties and standard techniques does not necessarily signal an end for the algorithm.  However, if those standard techniques can be combined in such a way that their overall structure is the same as the DCA, but their overall performance is better, then there is a danger that ``fixing'' the DCA will eradicate it entirely. Before clear guidance can be formulated on when, if ever, the DCA is an appropriate choice for a given application, it is important to explore all of its algorithmically unique components.  With the classification and filtering properties investigated the next properties that should come under scrutiny are the use of multiple timescales across the cell population and the sampling of antigen.

\bibliographystyle{plain}
\bibliography{mybib}

\end{document}